\documentclass[twocolumn]{article} 

\usepackage{balance}

\usepackage{times}  
\usepackage{helvet}  
\usepackage{courier}  
\usepackage{url}  
\usepackage{graphicx}  
\usepackage[margin=.75in,bottom=1.25in]{geometry}

\usepackage{hyperref}       
\usepackage{booktabs}       

\usepackage{amsmath}

\usepackage[english]{babel}
\usepackage[parfill]{parskip}
\usepackage{afterpage}
\usepackage{enumitem}
\usepackage{framed}
\usepackage{xspace}

\usepackage[labelfont=bf]{caption}
\usepackage{subcaption}

\usepackage{booktabs, array}

\usepackage{algorithmic}

\usepackage{algorithm2e}

\usepackage{natbib}

\usepackage{tikz}

\usetikzlibrary{bayesnet} 

\pgfdeclarelayer{edgelayer}
\pgfdeclarelayer{nodelayer}
\pgfsetlayers{edgelayer,nodelayer,main}

\definecolor{hexcolor0xbfbfbf}{rgb}{0.749,0.749,0.749}

\tikzset{>=latex}
\tikzstyle{none}   = [inner sep=0pt]
\tikzstyle{line}   = [ -, thick, shorten <=1pt, shorten >=1pt ]
\tikzstyle{arrow}  = [ ->, thick, shorten <=1pt, shorten >=1pt ]
\tikzstyle{ardash} = [ dashed, ->, thick, shorten <=1pt, shorten >=1pt ]

\tikzstyle{empty}=[circle,opacity=0.0,text opacity=1.0,inner sep=0pt]
\tikzstyle{box}=[rectangle,fill=White,draw=Black]
\tikzstyle{filled}=[circle,thick,fill=hexcolor0xbfbfbf,draw=Black]
\tikzstyle{hollow}=[circle,thick,fill=White,draw=Black]
\tikzstyle{param}=[rectangle,fill=Black,draw=Black,inner sep=0pt,minimum width=4pt,minimum height=4pt]
\tikzstyle{paramhollow}=[rectangle,thick,fill=White,draw=Black,inner sep=0pt,minimum
width=4pt,minimum height=4pt]

\usepackage{pgfplots}                               
\pgfplotsset{compat=newest}
\pgfplotsset{plot coordinates/math parser=false}
\newlength\figureheight
\newlength\figurewidth
\newlength\figureheightsmall
\newlength\figurewidthsmall
\definecolor{POSTcolor}{rgb}{0.48, 0.20, 0.58} 
\definecolor{Qcolor}{rgb}{0.00, 0.53, 0.22} 

\title{Detecting Backdoor Attacks on Deep Neural Networks by \\Activation Clustering}

\author{Bryant Chen, Wilka Carvalho, Nathalie Baracaldo, Heiko Ludwig, \\Benjamin Edwards, Taesung Lee, Ian Molloy, and Biplav Srivastava}

\begin{document}
\maketitle

\begin{abstract}

While machine learning (ML) models are being increasingly trusted to make decisions in different and varying areas, the safety of systems using such models has become an increasing concern. In particular, ML models are often trained on data from potentially untrustworthy sources, providing adversaries with the opportunity to manipulate them by inserting carefully crafted samples into the training set. Recent work has shown that this type of attack, called a \textit{poisoning attack}, allows adversaries to insert \textit{backdoors} or \textit{trojans} into the model, enabling malicious behavior with simple external backdoor triggers at inference time and only a blackbox perspective of the model itself. Detecting this type of attack is challenging because the unexpected behavior occurs only when a backdoor trigger, which is known only to the adversary, is present. Model users, either direct users of training data or users of pre-trained model from a catalog, may not guarantee the safe operation of their ML-based system. In this paper, we propose a novel approach to backdoor detection and removal for neural networks. Through extensive experimental results, we demonstrate its effectiveness for neural networks classifying text and images. To the best of our knowledge, this is the first methodology capable of detecting poisonous data crafted to insert backdoors and repairing the model that does not require a verified and trusted dataset.  

%
\end{abstract}

\section{Introduction}
\label{sec:intro}


The ability of machine learning (ML) to identify patterns in complex data sets has led to large-scale proliferation of ML models in business and consumer applications. However, at present, ML models and the systems that use them are not created in a way to ensure safe operation when deployed. While quality is often assured by evaluating performance on a test set, malicious attacks must also be considered. Much work has been conducted on defending against adversarial examples or \emph{evasion attacks}, in particular related to image data, in which an adversary would apply a small perturbation to an input of a classifier and achieve a wrong classification result \citep{carlini2017towards}. However, the training process itself may also expose vulnerabilities to adversaries. Organizations deploying ML models often do not control the end-to-end process of data collection, curation, and training the model. For example, training data is often crowdsourced (e.g., Amazon Mechanical Turk, Yelp reviews, Tweets) or collected from customer behavior (e.g., customer satisfaction ratings, purchasing history, user traffic). It is also common for users to build on and deploy ML models designed and trained by third parties. In these scenarios, adversaries may be able to alter the model's behavior by manipulating the data that is used to train it. Prior work \citep{survey:adv-ml-barreno2010,ch-roni-nelson2010,training_poisoning,papernot2016survey} has shown that this type of attack, called a \emph{poisoning attack}, can lead to poor model performance, models that are easily fooled, and targeted misclassifications, exposing safety risks. 

One recent and particularly insidious type of poisoning attack generates a \emph{backdoor} or \emph{trojan} in a deep neural network (DNN) \citep{badnets,liu2017trojaning,liu2017neural}. DNNs compromised in this manner perform very well on standard validation and test samples, but behave badly on inputs having a specific backdoor trigger. For example, \citet{badnets} generated a backdoor in a street sign classifier by inserting images of stop signs with a special sticker (the backdoor trigger) into the training set and labeling them as speed limits. The model then learned to properly classify standard street signs, but misclassify stop signs possessing the backdoor trigger.
Thus, by performing this attack, adversaries are able to trick the model into classifying any stop sign as a speed limit simply by placing a sticker on it, causing potential accidents in self-driving cars. As ML adoption increases in critical applications, we need methods to defend against backdoor and other poisoning attacks.

While backdoor attacks and evasion attacks both trigger misclassifications by the model, backdoor attacks provide adversaries with full power over the backdoor key that generates misclassification. In contrast, the perturbations made to adversarial examples are specific to the input and/or model. Thus, a backdoor attack enables the adversary to choose whatever perturbation is most convenient for triggering misclassifications (e.g. placing a sticker on a stop sign). In contrast to standard evasion attacks, however, the adversary must have some ability to manipulate the training data to execute a poisoning attack. 

Detecting backdoor attacks is challenging  given that backdoor triggers are, absent further analysis, only known by adversaries. Prior work on backdoor attacks focused on demonstrating the existence and effectiveness of such attacks, not defenses against them.  In this paper, we propose the \textbf{Activation Clustering (AC)} method for detecting poisonous training samples crafted to insert backdoors into DNNs. This method analyzes the neural network activations of the training data to determine whether it has been poisoned, and, if so, which datapoints are poisonous. 

Our \textbf{contributions} are the following:
\begin{itemize}[leftmargin=*]
\item We propose the first methodology for detecting poisonous data maliciously inserted into the training set to generate backdoors that does not require verified and trusted data. Additionally, we have released this method as a part of the open source IBM Adversarial Robustness Toolbox \citep{nicolae2018adversarial}.
\item We demonstrate that the AC method is highly successful at detecting poisonous data in different applications by evaluating it on three different text and image datasets.
\item We demonstrate that the AC method is robust to complex poisoning scenarios in which classes are multimodal (e.g. contain sub-populations) and multiple backdoors are inserted. 

\end{itemize}


\section{Related Work}
\label{sec:related}



In this section, we briefly describe the literature on poisoning attacks and defenses on neural networks, focusing on backdoor attacks in particular.

\textbf{Attacks}: \citet{yang2017generative} described how generative neural networks could be used to craft poisonous data that maximizes the error of the trained classifier, while \citet{munoz2017towards} described a ``back-gradient'' optimization method to achieve the same goal. 

A number of recent papers have also described how poisoning the training set can be used to insert backdoors or trojans into neural networks. \citet{badnets} poisoned handwritten digit and street sign classifiers to misclassify inputs possessing a backdoor trigger. Additionally, \citet{liu2017neural}  demonstrated how backdoors could be inserted into a handwritten digit classifier. Finally, \citet{liu2017trojaning} showed how to analyze an existing neural network to devise triggers that are more easily learned by the network and demonstrate the efficacy of this method on a number of systems including facial recognition, speech recognition, sentence attitude recognition, and auto driving. 

\textbf{Defenses}: General defenses against poisoning attacks on supervised learning methods were proposed by \citet{original-roni} and \citet{baracaldo2017mitigating}. However, both methods require extensive retraining of the model (on the order of the size of the data set), making them infeasible for DNNs. Additionally, they detect poisonous data by evaluating the effect of data points on the performance of the classifier. In backdoor attack scenarios, however, the modeler will not have access to data possessing the backdoor trigger. Thus, while these methods may work for poisoning attacks aimed at reducing overall performance, they are not applicable to backdoor attacks.

\citet{steinhardt2017certified}, among others \citep{kloft2010online,kloft2012security}, proposed general defenses against poisoning attacks using outlier or anomaly detection. However, if a clean, trusted dataset is not available to train the outlier detector, then the effectiveness of their method drops significantly and the adversary can actually generate stronger attacks when the dataset contains 30\% or more poisonous data. Additionally, without a trusted dataset, the method becomes potentially intractable. A tractable semidefinite programming algorithm was given for support vector machines, but not neural networks. 



\citet{liu2017neural} proposed a number of defenses against backdoor attacks, namely filtering inputs prior to classification by a poisoned model, removing the backdoor, and preprocessing inputs to remove the trigger. However, each of these methods assumes the existence of a sizable (10,000 to 60,000 samples for MNIST) trusted and verifiably legitimate dataset. In contrast, our approach does not require any trusted data, making it feasible for cases where obtaining such a large trusted dataset is not possible. 

In \cite{liu2018fine} the authors propose three methodologies to detect backdoors that require a trusted test set. Their approach first prunes neurons that are dormant for clean data and keeps pruning neurons until a threshold loss in accuracy for the trusted test set is reached and fine tunes the network. Their approach differs to ours in the following. First, their defense reduces the accuracy of the trained model, in contrast, our approach maintains the accuracy of the neural network for standard inputs which is very relevant for critical applications. Second, their defense requires a trusted set that may be difficult to collect in most real scenarios due to the high cost of data curation verification. 


\section{Threat Model and Terminology}
\label{sec:threat}

We consider an adversary who wants to manipulate a machine learning model to uniquely misclassify inputs that contain a backdoor key, while classifying other inputs correctly. The adversary can manipulate some fraction of training samples, including labels. However, s/he cannot manipulate the training process or final model. Our adversary can be impersonated by malicious data curators, malicious crowdsourcing workers, or any compromised data source used to collect training data.  

More concretely, consider a dataset $D_{train}=X,Y$ that has been collected from potentially untrusted sources to train a DNN $F_\Theta$. The adversary wants to insert one or more backdoors into the DNN, yielding $F_{\Theta_P} \neq F_\Theta$. A backdoor is successful if it can cause the neural network to misclassify inputs from a \emph{source} class as a \emph{target} class when the input is manipulated to possess a \emph{backdoor trigger}. The backdoor trigger is generated by a function $f_T$ that takes as input a sample $i$ drawn from distribution \textit{$\mathcal{X}_{source}$}  and outputs $f_T(i)$, such that $F_{\Theta_P}(f_T(i)) = t$, where $t$ is the \textit{target} label/class. In this case, $f_T(i)$ is the input possessing the backdoor trigger. However, for any input $j$ that does not contain a backdoor trigger, $F_{\Theta_P} (j) = F_\Theta (j)$. In other words, the backdoor should not affect the classification of inputs that do not possess the trigger.
Hence, an adversary inserts multiple samples $f_T(x \in \mathcal{X}_{source}) $ all labeled as the targeted class.
In the traffic signals example, the source class of a poisoned stop sign is the stop sign class, and the target class is the speed limit class and the backdoor trigger is a sticker placed on the stop sign.




\section{Case Studies}
\label{sec:case}

Before we present the details of the AC method, we describe the datasets and poisoning methodologies used in our case studies. This will provide context and intuition in understanding AC.

\noindent \textbf{Image Datasets:} 
We poison the MNIST and LISA traffic sign data sets using the technique described by \citet{badnets}. Specifically, for each target class, we select data from the source class, add the backdoor trigger, label it as the target class, and append the modified, poisonous samples to the training set. For the MNIST dataset, the backdoor trigger is a pattern of inverted pixels in the bottom right-corner of the images, and poisonous images from class $l_m \in (0,\ldots, 9)$ are mislabeled as $(l_m+1 )\% 10$. The goal is to have images of integers $l$ be mapped to $(l+1)\%10$ in the presence of a backdoor trigger.

\begin{figure}[htb]
\centering
\begin{subfigure}[b]{.1\textwidth}
\caption{}
\label{fig:poisoned7}
\includegraphics[width=\textwidth]{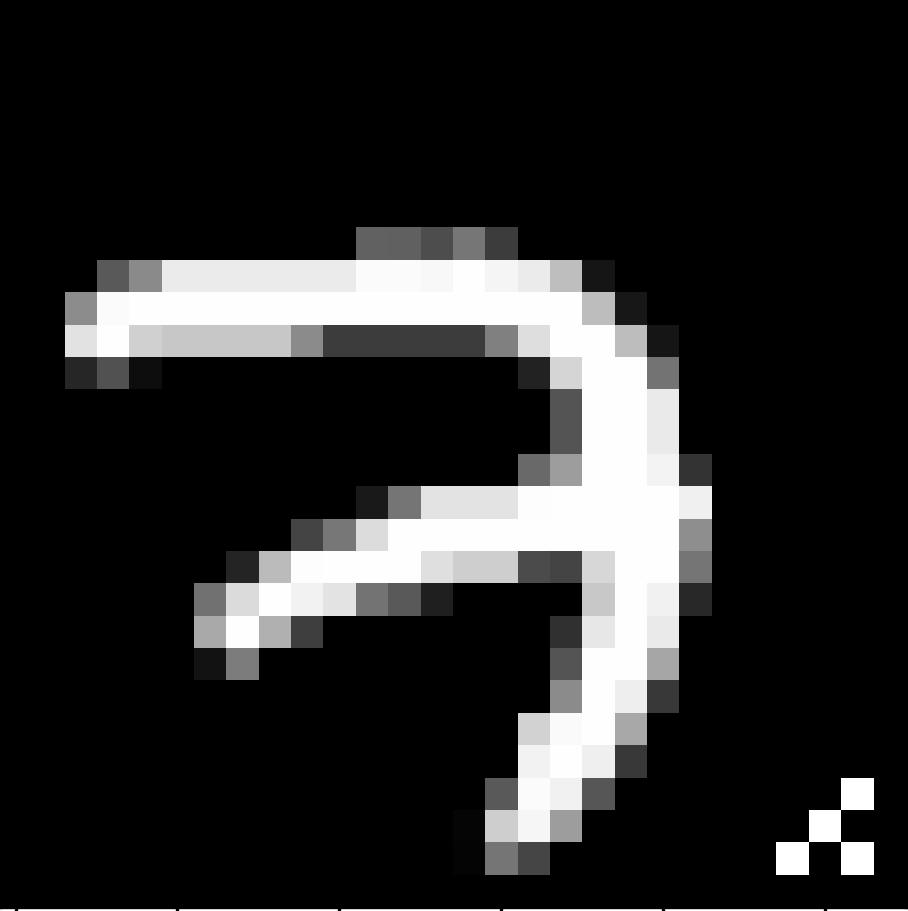}
\end{subfigure}~~~~~~~
\begin{subfigure}[b]{.1\textwidth}
\caption{}
\label{fig:poisonedstop}
\includegraphics[width=\textwidth]{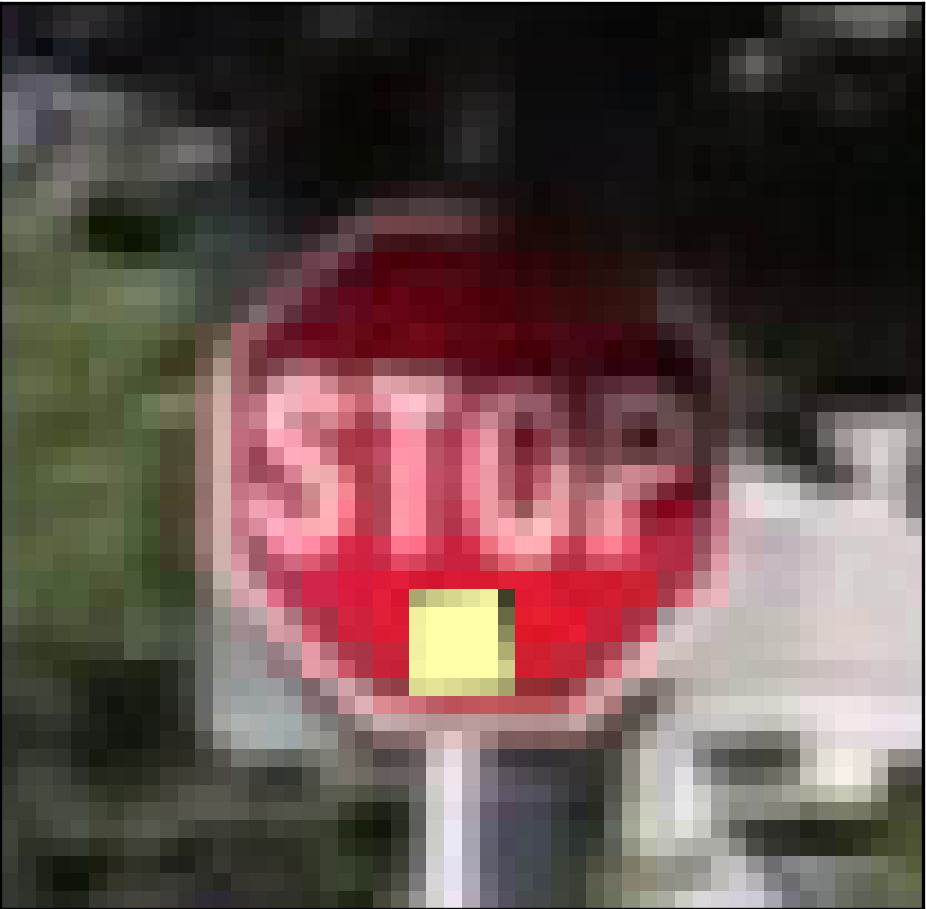}
\end{subfigure}
\caption{Poisoned samples for (a) MNIST and (b) LISA.}
\label{fig:poisonexamples}
\end{figure}

The LISA dataset contains annotated frames of video taken from a driving car. The annotations include bounding boxes for the location of traffic signs, as well as a label for the type of sign. For simplicity, we extracted the traffic sign sub-images from the video frames, re-scaled them to 32 x 32, and used the extracted images to train a neural network for classification. Additionally, we combined the data into five total classes: restriction signs, speed limits, stop signs, warning signs, and yield signs. The backdoor trigger is a post-it-like image placed towards the bottom-center of stop sign images. Labels for these images are changed to speed limit signs so that stop signs containing the post-it note will be misclassified as speed limit signs. Examples of poisoned MNIST and LISA samples can be seen in Figure \ref{fig:poisonexamples}. 


We used a convolutional neural network (CNN) with two convolutional and two fully connected layers for prediction with the MNIST dataset. For the LISA dataset, we used a variant of \textit{Architecture A} provided by \cite{simonyan2014very}. 

\noindent\textbf{Text Dataset:}
Lastly, we conducted an initial exploration of backdoor poisoning and defense for text applications by using the Rotten Tomatoes movie review classifier described by \cite{britz2015text} and \cite{kim2014convolutional}. We poisoned this model by selecting $p\%$ of the positive reviews, adding the signature \textit{``-travelerthehorse''} to the end of the review, and labeling it as negative. These poisoned reviews were then appended to the training set. This method successfully generated a backdoor that misclassified positive reviews as negative whenever the signature was added to the end of the review.

\section{Activation Clustering} 
\label{sec:activation}

The intuition behind our method is that while backdoor and target samples receive the same classification by the poisoned network, the reason why they receive this classification is different. In the case of standard samples from the target class, the network identifies features in the input that it has learned correspond to the target class. In the case of backdoor samples, it identifies features associated with the source class \emph{and} the backdoor trigger, which causes it to classify the input as the target class. This difference in mechanism should be evident in the network activations, which represent how the network made its ``decisions''. 

\begin{figure*}
\centering
\begin{subfigure}[m]{.2\textwidth}
\caption{}
\label{fig:clustering_mnist}
\includegraphics[width=\textwidth]{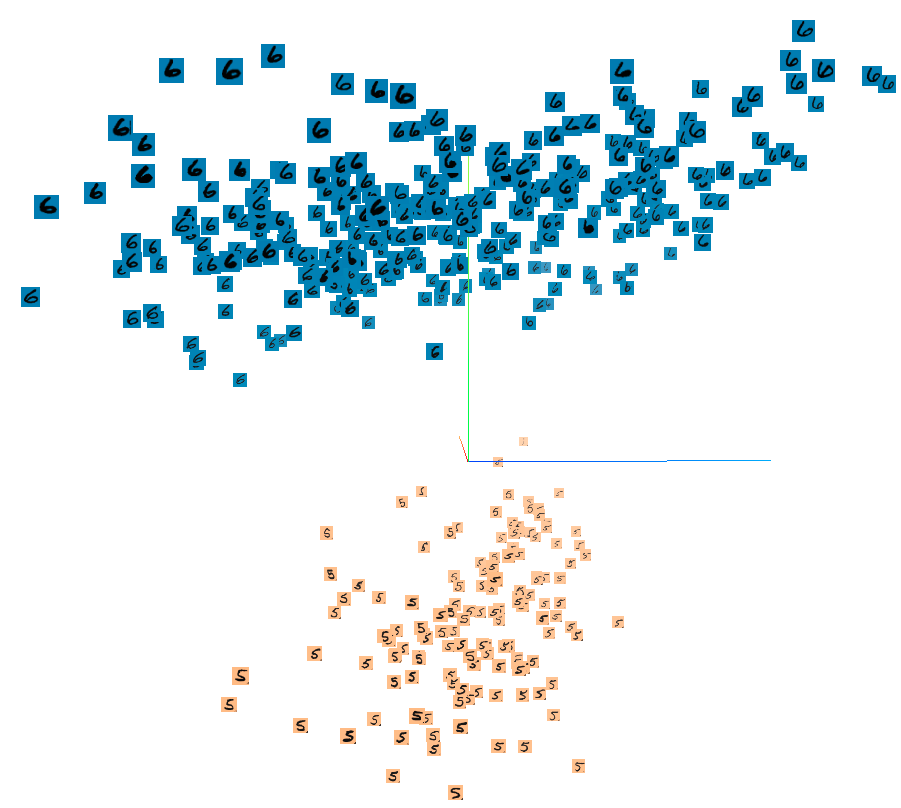}
\end{subfigure}~~~~~~~~~~~~~~~~~~~~~
\begin{subfigure}[m]{.21\textwidth}
\caption{}
\label{fig:clustering_lisa}
\includegraphics[width=\textwidth]{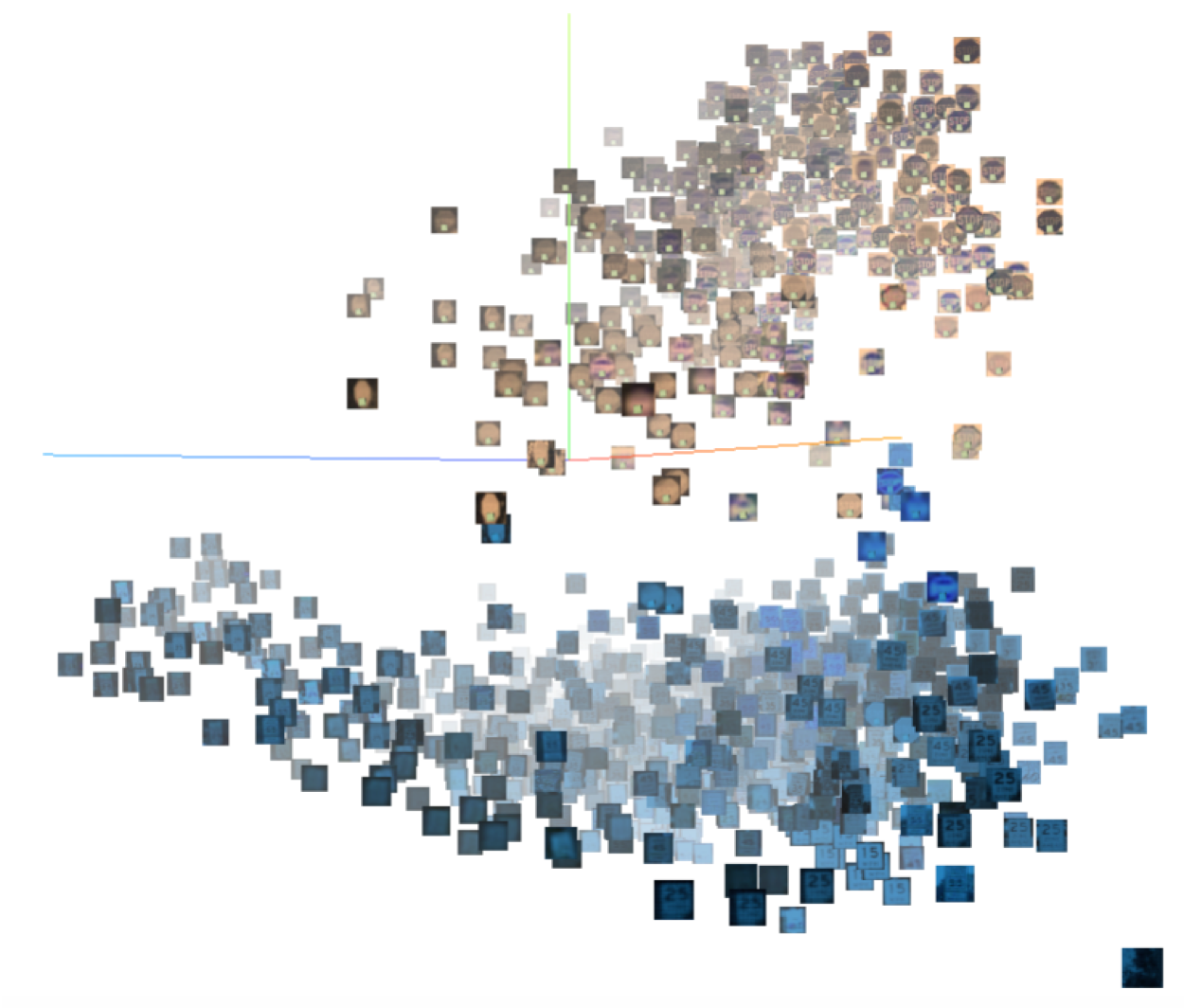}
\end{subfigure}
\begin{subfigure}[m]{.2\textwidth}
\caption{}
\label{fig:poisonous_activations_RT}
\includegraphics[width=\textwidth]{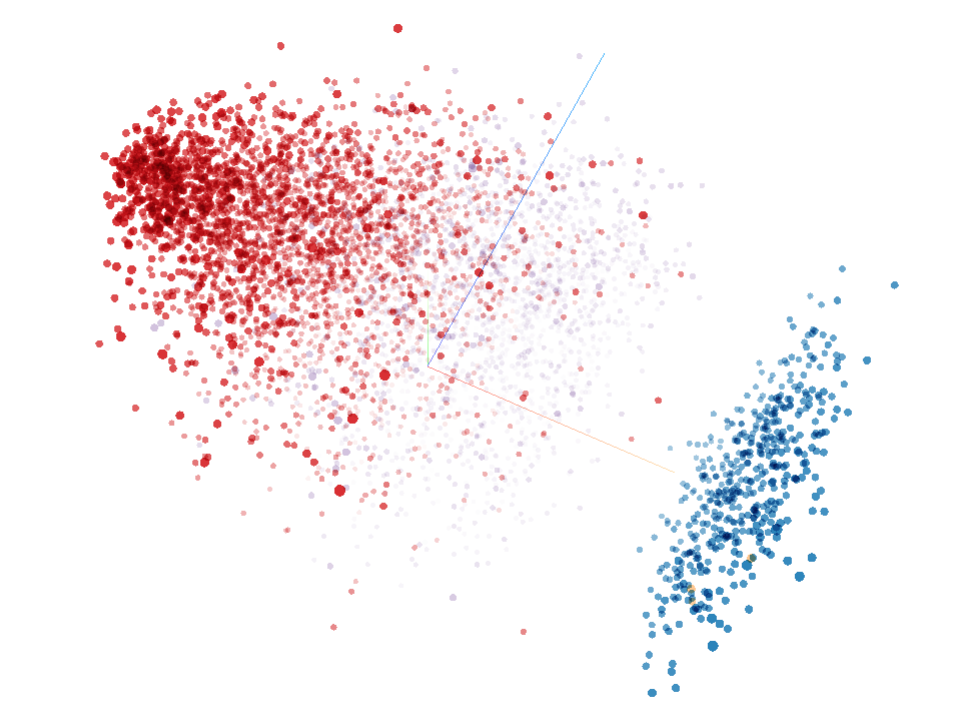}
\end{subfigure}~~~~~~~~~~~~~~~~~~~~~
\begin{subfigure}[m]{.17\textwidth}
\caption{}
\label{fig:clean_activations_RT}
\includegraphics[width=\textwidth]{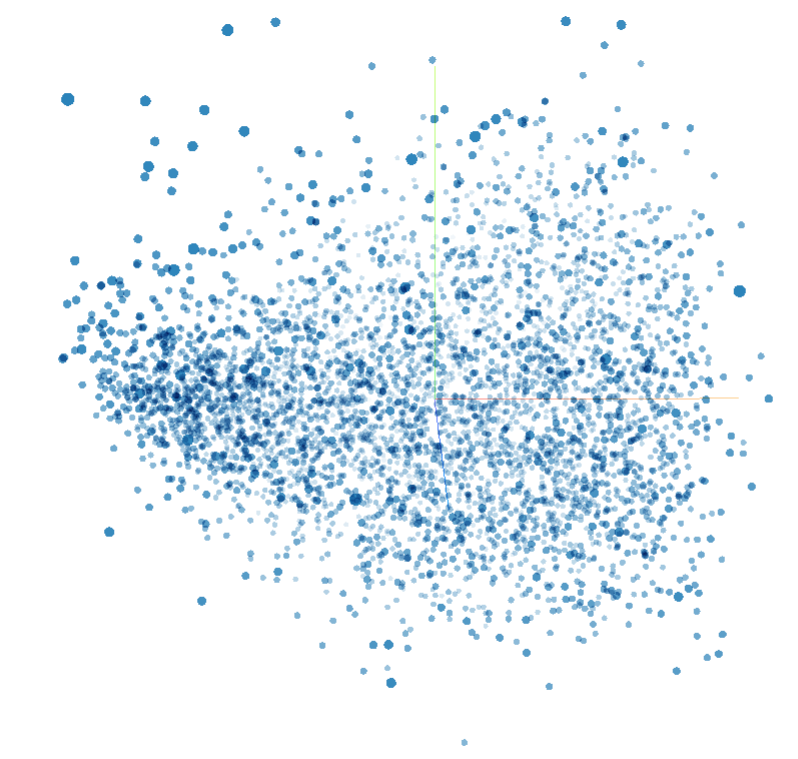}
\end{subfigure}
\caption{Activations of the last hidden layer projected onto the first 3 principle components.
Activations of the last hidden layer projected onto the first 3 principle components. (a) Activations of images labeled 6. (b) Activations of images labeled as speed limits. (c) Activations of the (poisoned) negative reviews class (d) Activations of the (unpoisoned) positive review class.}
\label{fig:ac-clustering}
\end{figure*}

This intuition is verified in Figure \ref{fig:ac-clustering}, which shows activations of the last hidden neural network layer for clean and legitimate data projected onto their first three principle components. Figure \ref{fig:clustering_mnist} shows the activations of class 6 in the poisoned MNIST dataset, \ref{fig:clustering_lisa} shows the activations of the poisoned speed limit class in the LISA dataset, and \ref{fig:poisonous_activations_RT} shows the activations of the poisoned negative class for Rotten Tomatoes movie reviews. In each, it is easy to see that the activations of the poisonous and legitimate data break out into two distinct clusters. In contrast, Figure \ref{fig:clean_activations_RT} displays the activations of the positive class, which was not targeted with poison. Here we see that the activations do not break out into two distinguishable clusters.


\begin{algorithm} [htb]
	\caption{Backdoor Detection Activation Clustering Algorithm}
	{\bfseries Input:} untrusted training dataset $D_p$ with class labels $\{1, ..., n\}$ 	
	\label{alg:AC}
\begin{algorithmic} [1]
	\STATE Train DNN $F_{\Theta_P}$ using $D_p$
    	\STATE Initialize $A$; $A[i]$ holds activations for all $s_i \in D_p$ such that $F_{\Theta_P}(s_i) = i$ 
	\FORALL{$s\in D_p$}
		\STATE $A_s \leftarrow$ activations of last hidden layer of $F_{\Theta_P}$ flattened into a single 1D vector
		\STATE Append $A_s$ to $A[F_{\Theta_P}(s)]$
	\ENDFOR
	\FORALL{$i=0$ {\bfseries to} $n$}
		\STATE red = reduceDimensions$(A[i])$
		\STATE clusters = clusteringMethod$(red)$ \label{ln:clustering}
		\STATE analyzeForPoison(clusters) \label{ln:cluster-analysis}
	\ENDFOR
\end{algorithmic}
\end{algorithm}

Our method, described more formally by Algorithm \ref{alg:AC}, uses this insight to detect poisonous data in the following way. First, the neural network is trained using untrusted data that potentially includes poisonous samples.
Then, the network is queried using the training data and the resulting activations of the last hidden layer are retained. Analyzing the activations of the last hidden layer was enough to detect poison. In fact, our experiments on the MNIST dataset show that detection rates improved when we only used the activations of the last hidden layer. Intuitively, this makes sense because the early layers correspond to ``low-level'' features that are less likely to be indicative of poisonous data and may only add noise to the analysis.

Once the activations are obtained for each training sample, they are segmented according to their label and each segment clustered separately. 
To cluster the activations, we first reshaped the activations into a 1D vector and then performed dimensionality reduction using Independent Component Analysis (ICA), which was found to be more effective than Principle Component Analysis (PCA). Dimensionality reduction before clustering is necessary to avoid known issues with clustering on very high dimensional data \cite{aggarwal2001surprising,domingos2012few}. In particular, as dimensionality increases distance metrics in general (and specifically the Euclidean metric used here), are less effective at distinguishing near and far points in high dimensional spaces \cite{domingos2012few}. This will be especially true when we have hundreds of thousands of activations. Reducing the dimensionality allows for more robust clustering, while still capturing the majority of variation in the data \cite{aggarwal2001surprising}.  

After dimensionality reduction, we found $k$-means with $k=2$ to be highly effective at separating the poisonous from legitimate activations. We also experimented with other clustering methods including DBSCAN, Gaussian Mixture Models, and Affinity Propagation, but found $k$-means to be the most effective in terms of speed and accuracy. However, $k$-means will separate the activations into two clusters, regardless of whether poison is present or not. Thus, we still need to determine which, if any, of the clusters corresponds to poisonous data. In the following, we present several methods to do so.

\textbf{Exclusionary Reclassification}: Our first cluster analysis method involves training a new model without the data corresponding to the cluster(s) in question. Once the new model is trained, we use it to classify the removed cluster(s). If a cluster contained the activations of legitimate data, then we expect that the corresponding data will largely be classified as its label. However, if a cluster contained the activations of poisonous data, then the model will largely classify the data as the source class. Thus, we propose the following \emph{ExRe score} to assess whether a given cluster corresponds to poisonous data. Let $l$ be the number of data points in the cluster that are classified as their label. Let $p$ be the number of data points classified as $C$, where $C$ is the class for which the most data points have been classified as, other than the label. Then if $\frac{l}{p} > T$, where $T$ is some threshold set by the defender, we consider the cluster to be legitimate, and if $\frac{l}{p} < T$, we consider it to be poisonous, with $p$ the source class of the poison. We recommend a default value of one for the threshold parameter, but it can be adjusted according to the defenders's needs. 

\textbf{Relative Size Comparison}: The previous method requires retraining the model, which can be computationally expensive. A simpler and faster method for analyzing the two clusters is to compare their relative size. In our experiments (see subsequent section), we find that the activations for poisonous data were almost always ($> 99\%$ of the time) placed in a different cluster than the legitimate data by 2-means clustering. Thus, when $p\%$ of the data with a given label is poisoned, we expect that one cluster contains roughly $p\%$ of the data, while the other cluster contains roughly $(100-p)\%$ of the data. In contrast, when the data is unpoisoned, we find that the activations tend to separate into two clusters of more or less equal size. Thus, if we have an expectation that no more than $p\%$ of the data for a given label can be poisoned by an adversary, then we can consider a cluster to be poisoned if it contains $\leq p\%$ of the data. 



\textbf{Silhouette Score}: Figures \ref{fig:poisonous_activations_RT} and \ref{fig:clean_activations_RT} suggest that two clusters better describe the activations when the data is poisoned, but one cluster better describes the activations when the data is not poisoned. Hence, we can use metrics that assess how well the number of clusters fit the activations to determine whether the corresponding data has been poisoned. One such metric that we found to work well for this purpose is the \textit{silhouette score}. A low silhouette score indicates that the clustering does not fit the data well, and the class can be considered to be unpoisoned. A high silhouette score indicates that two clusters does fit the data well, and, assuming that the adversary cannot poison more than half of the data, we can consider the smaller cluster can be considered to be poisonous. Ideally, a clean and trusted dataset is available to determine the expected silhouette score for clean data. Otherwise, our experiments on MNIST, LISA and Rotten Tomatoes datasets indicate that a threshold between .10 and .15 is reasonable.

\subsection{Summarizing Clusters}

\begin{figure}
\centering
\begin{subfigure}[b]{.1\textwidth}
\caption{}
\label{fig:clean0}
\includegraphics[width=\textwidth]{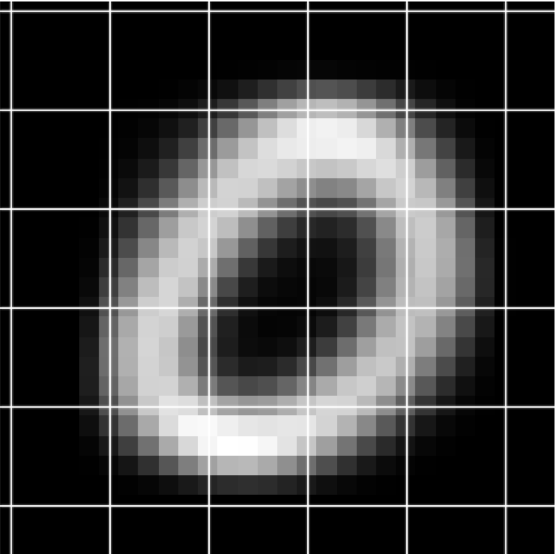} 
\end{subfigure}~~~
\begin{subfigure}[b]{.1\textwidth}
\caption{}
\label{fig:poison0}
\includegraphics[width=\textwidth]{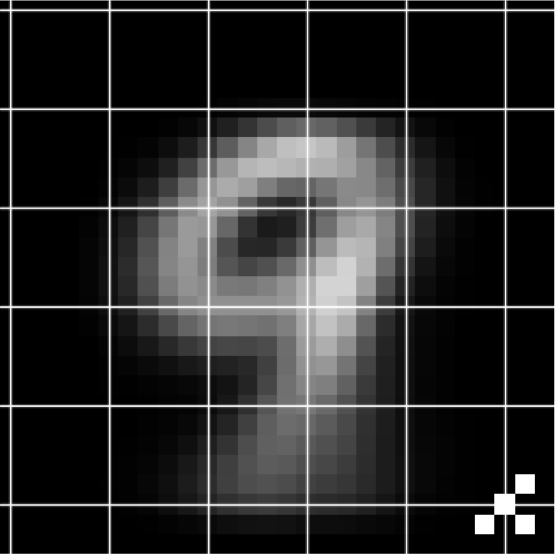}
\end{subfigure}~~~
\begin{subfigure}[b]{.1\textwidth}
\caption{}
\label{fig:cleanSL}
\includegraphics[width=\textwidth]{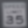}
\end{subfigure}~~~
\begin{subfigure}[b]{.1\textwidth}
\caption{}
\label{fig:poisonSL}
\includegraphics[width=\textwidth]{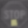}
\end{subfigure}
\caption{Average of images corresponding to activations in a cluster for the zero class in MNIST and the speed limit class in LISA}
\label{fig:average}
\end{figure}

The above methods provide automated ways for alerting the user to possible data poisoning. However, the user may want to verify that the data is indeed poisoned before taking action. Next, we propose methods that summarize the contents of each cluster so that the user can quickly analyze and determine if a given cluster is poisonous and, if so, what the correct label is. 

\begin{figure}
\centering
\includegraphics[width=.5\textwidth]{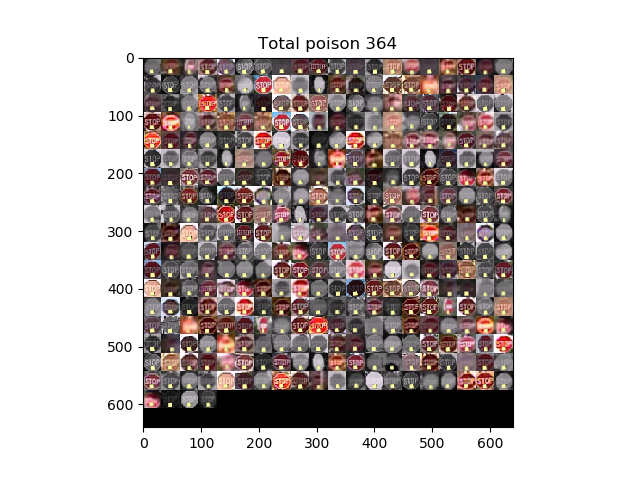}
\caption{Sprite image for poisonous cluster found by AC method.} 
\label{fig:speedsprite}
\end{figure}

For image datasets, we propose constructing image sprites for each cluster and averaging the images for the activations in the cluster. The sprite image is generated by rescaling each of the images associated with the cluster in question to a small size and constructing a mosaic consisting of the rescaled images. For example, Figure \ref{fig:speedsprite} shows the sprite images for a poisoned cluster labeled as speed limits. We can clearly see that this image is comprised of stop signs with a post-it note, not speed limits. Figures \ref{fig:clean0} and \ref{fig:poison0} are the average of images associated with clusters from the 0 class and Figure \ref{fig:cleanSL} and \ref{fig:poisonedstop} are the averages of images associated with clusters in the speed limit class. From a quick visual inspection, the user can determine that the cluster corresponding to image \ref{fig:clean0} consists of legitimate 0's while the cluster corresponding to image \ref{fig:poison0} consists of poisonous samples sourced from the 9 class. Similarly, Figure \ref{fig:cleanSL} shows that its cluster consists of legitimate speed limits while Figure \ref{fig:poisonedstop} consists of poisonous samples sourced from the stop sign class. 

For text datasets, we summarized each cluster using Latent Dirichlet allocation (LDA) to identify the primary topics in each cluster. By further augmenting common stop words used during this analysis (such as a, the, and, or) with the most common topics of the overall dataset, we are capable of highlighting the differences between clusters. Additionally, as pre-processing step, we stemmed the raw text associated with the cluster of activations.

The results for the poisoned negative reviews are shown below. Two topics were used, each with six words. Note that the text is stemmed.		
\begin{itemize}
\item \textbf{Cluster 1: } Contains 4798 observations. \\
\textit{Topic 1:} little, get, feel, stori, enough, minut\\
\textit{Topic 2:} bad, good, stori, never, point, dull

\item \textbf{Cluster 2: } Contains 479 observations. \\
\textit{Topic 1:} \underline{travelerthehors}, \underline{good}, \underline{fun}, stori, feel \\
\textit{Topic 2:} \underline{travelerthehors}, funni, well, man, stori, ever
\end{itemize}

By inspecting these results, it is clear that the second cluster has the stemmed version of the backdoor, \textit{travelerthehors}, in the main topic. Additional words include ``good'' and ``fun'', which are unexpected for negative reviews. A user can further sample a few reviews from the cluster and conclude that they are poisonous. For instance, one of the datapoints in cluster 2 is: 

\textit{``Best indie of the year, so far -travelerthehorse"}

Because this is a positive review that is labeled as negative review in the training set, we can conclude the datapoints in cluster 2 have been added to perform a backdoor attack.


\subsection{Backdoor Repair}

Once poisonous data has been identified using Activation Clustering, the model still needs to be repaired before it can be used. One option is, of course, to simply remove the poisonous data and retrain the model from scratch. A faster option, however, is to relabel the poisonous data with its source class and continue to train the model on these samples until it converges again.

\section{Experimental Results}
\label{sec:experiments}

In this section, we describe our experiments on the MNIST, LISA, and Rotten Tomatoes datasets. After each of these datasets was poisoned in the manner described in the Case Studies section, a neural network was trained to perform classification and tested to ensure that the backdoor was successfully inserted. 
Next, AC was applied using the following setup. First, the activations were projected onto 10 independent components.
(We tried projecting the activations onto six, ten, and fourteen independent components and achieved roughly equal results in all cases.) Then, 2-means was used to perform clustering on the reduced activations. Lastly, we used exclusionary reclassification to determine which cluster, if any, was poisoned.

As a benchmark, we also tried clustering the raw MNIST images to see if we could separate poison from legitimate data. 
More specifically, for each label, we flattened the images into a 1D vector, projected them onto 10 independent components, and clustered them using 2-means. Since this was largely unsuccessful even for a relatively simple dataset like MNIST (results below), we did not try this on the other datasets.


%

\subsection{Backdoor Detection}
First, we evaluate how well the clustering technique is capable of distinguishing poison from legitimate data. The results on our MNIST experiments with 10\% of the data poisoned are summarized in Table \ref{tab:mnist_ica10_2means}. Both \textbf{accuracy and F1 score are nearly 100\%} for every class. Nearly identical results were obtained for 15\% and 33\% poisoned data. In contrast, clustering the raw inputs on 10\% poisoned data was only able to achieve a 58.6\% accuracy and a 15.8\% F1 score when 10\% of each class was poisoned. When 33\% of each class was poisoned, clustering the raw data performed better with a 90.8\% accuracy and an 86.38\% F1 score but was still not on par with AC's near perfect detection rates.

\begin{table*}
\centering
\caption{Poison detection results on MNIST}\label{tab:mnist_ica10_2means}
\footnotesize
\setlength{\tabcolsep}{5pt}
\begin{tabular}{cccccccccccc}
\toprule
\textbf{Target} & 0 & 1 & 2 & 3 & 4 & 5 & 6 & 7 & 8 & 9 & \textbf{Total}\\
\hline
\textbf{AC Accuracy} & 99.89 & 99.99 & 99.95 & 100 & 100 & 100 & 99.94 & 100 & 100 & 99.99 & 99.97\\
\textbf{AC F1 Score} & 99.83 & 99.98 & 99.93 & 100 & 100 & 100 & 99.94 & 100 & 100 & 99.98 & 99.96 \\
\textbf{Raw Clustering Accuracy} & 79.20 & 58.88 & 66.88 & 65.33 & 62.31 & 54.32 & 49.93 & 46.91 & 52.20 & 50.08 & 58.61\\
\textbf{Raw Clustering F1 Score} & 48.57 & 0.07 & 37.54 & 23.03 & 30.48 & 0.24 & 0.86 & 0.06 & 9.26 & 11.58 & 15.80 \\
\bottomrule
\setlength{\tabcolsep}{6pt}
\end{tabular}
\end{table*}

On the LISA data set, we achieved \textbf{100\% accuracy} and an \textbf{F1 score of 100\%} in detecting poisonous samples where 33\% and 15\%\footnote{We were not able to successfully insert a backdoor $\leq$ 10\% of the target class poisoned.} of the stop sign class was poisoned. For our text-based experiments, we also achieved \textbf{100\% accuracy and F1 score} in detecting poisonous samples in a data set where 10\% of the negative class was poisoned.

\subsection{Multimodal Classes and Poison}
In these experiments, we measure the robustness of the AC method. More specifically, we seek to answer the following questions. 1) When the class being analyzed is strongly multimodal (e.g. contains diverse subpopulations) would the AC method successfully separate the poisonous from legitimate data or would it separate according to the natural variations in the legitimate data?
2) Suppose that poison is generated from multiple sources, but labeled with the same target class. Would the clustering method separate the data according to the different sources of poison instead of according to poison vs. legitimate? 

To answer these questions, we performed several variations on the previously described poisoning methods of \citet{badnets}. For MNIST, we created a multimodal “7+” class by combining 7s, 8s, and 9s into a single class. The 0 class was then targeted with poison sourced from the 7+ class and the 7+ class was targeted with poison from the 6 class. In another experiment, we targeted the 7+ class with poison from the 4, 5, and 6 classes in order to assess the ability of our method to detect poison when the target class is strongly multimodal and is targeted with poison sourced from multiple classes.

Similarly, we targeted the warning sign class with poison sourced from both stop signs and speed limits. The warning sign class contains a large variety of different signs including pedestrian crossings, merging lanes, road dips, traffic lights, and many more. Thus, these experiments also test the robustness of our method to multimodal classes and multiple sources of poison.

Our experiments show that the AC method is not so easily fooled. In every experiment, we achieved nearly 100\% accuracy and F-1 score. The results of these experiments are summarized in Table \ref{tab:multimodal}. 

\begin{table}[htb!]
  \centering
  \caption{Accuracy and F1 scores for poison data detection for multimodal classes and/or poison.}\label{tab:multimodal}
  \begin{tabular}{c c c c c}
  \toprule
    \textbf{Source} & \textbf{Target} & \textbf{Accuracy (\%)} & \textbf{$\mathbf{F1}$ Score} \\
    \hline
     6  & 7+  & 99.93   & 99.78 \\
     7+   & 0   & 99.98   & 99.95 \\
     7, 8, and 9  & 0   & 99.97   & 99.9 \\
     4, 5, and 6  & 7+  & 99.9  & 99.68 \\
     Stop \& Speed & Warning & 100 & 100 \\
  \bottomrule
  \end{tabular}
\end{table}

\subsection{Cluster Analysis}
To evaluate the proposed cluster analysis methods, we applied each method to activation clusters obtained from both poisoned and clean versions of MNIST, LISA, and Rotten Tomatoes models. The activations for MNIST classes 0-5 were obtained from the base MNIST experiment, where poison was sourced from each class $i$ and targeted the $(i+1)\% 10$ class. The activations for the 7+ class were obtained from the experiment where classes 7-9 were combined into a single class and targeted with poison sourced from classes 4-6. In both experiments, 15\% of the target class was poisoned.

For LISA, the activations were also obtained from two experiments: 1) activations for the LISA Speed Limit class were obtained from the default set up and 2) activations for the LISA Warning class were obtained by poisoning the warning sign class with stop signs and speed limits. 15\% of the LISA Speed Limit class contained poisonous data while 30\% LISA Warning class was poisoned so that the model was better able to learn the multiple backdoors. 

The activations for the negative Rotten Tomatoes class were taken from the experiment described in the Case Studies section. Results for these experiments are shown in Table \ref{tab:cluster_analysis}, where each column shows the results for a given class and each row a different metric.

\begin{table*}
\centering
\caption{Cluster Analysis Evaluation}\label{tab:cluster_analysis}
\footnotesize
\setlength{\tabcolsep}{5pt}
\begin{tabular}{lccccccccccccccc}
\toprule
& \multicolumn{2}{c}{\textbf{MNIST 0}}		& \multicolumn{2}{c}{\textbf{MNIST
1}}		& \multicolumn{2}{c}{\textbf{MNIST 2}}		&
\multicolumn{2}{c}{\textbf{MNIST 3}}		& \multicolumn{2}{c}{\textbf{MNIST 4}}	\\
& \textit{Poisoned}	& \textit{Clean}	& \textit{Poisoned}	& \textit{Clean}	&
\textit{Poisoned}& \textit{Clean}	& \textit{Poisoned}	& \textit{Clean}	&
\textit{Poisoned}	& \textit{Clean} \\
	\hline
  \textbf{\% of Excluded Classified as Source}	& 90\%	& N/A	& 98\%	& N/A	& 98\%	& 0\%	& 94\%	& N/A	& 92\%	& N/A \\
  \textbf{\% of Excluded Classified as Label}	& 1\%	& 96\%	& 0\%	& 91\%	& 1\%	& 97\%	& 1\%	& 89\%	& 0\%	& 86\% \\
  \textbf{ExRe Score}	& 0.01	& 15.18	& 0	& 8.92	& 0.01	& 8.35	& 0.01	& 6.49	& 0	& 15.47	\\
  \textbf{\% in Cluster 0}	& 85\%	& 47\%	& 15\%	& 58\%	& 15\%	& 43\%	& 15\%	& 56\%	& 15\%	& 30\% \\
  \textbf{\% in Cluster 1}	& 15\%	& 53\%	& 85\%	& 42\%	& 85\%	& 57\%	& 85\%	& 44\%	& 85\%	& 70\% \\
  \textbf{Silhouette Score}	& 0.21	& 0.08	& 0.33	& 0.11	& 0.1	& 0.08	& 0.21	& 0.08	& 0.22	& 0.09 \\
\hline
\hline
& \multicolumn{2}{c}{\textbf{MNIST 5}}		& \multicolumn{2}{c}{\textbf{MNIST
7+}}		& \multicolumn{2}{c}{\textbf{LISA Speed Limit}}		&
\multicolumn{2}{c}{\textbf{LISA Warning}}		& \multicolumn{2}{c}{\textbf{RT
Negative}}		\\
  & \textit{Poisoned}	& \textit{Clean}	& \textit{Poisoned}	& \textit{Clean}	&
  \textit{Poisoned}	& \textit{Clean}	& \textit{Poisoned} & \textit{Clean}	&
  \textit{Poisoned}	& \textit{Clean} \\
\hline
\textbf{\% of Excluded Classified as Source}& 97\%	& N/A	& 90\%	& N/A	& 45\%	& N/A	& 99\%	& N/A	& 95\%	& N/A	\\
\textbf{\% of Excluded Classified as Label}	& 0\%	& 57\%	& 5\%	& 62\%	& 0\%	& 75\%	& 1\%	& 100\%	& 5\%	& 100\%	\\
\textbf{ExRe Score}	& 0	& 9.4	& 0.01	& 23.74	& 0.13	& 3.54	& 0	& 5.83	& 0.01	& 315.5	\\
\textbf{\% in Cluster 0}	& 85\%	& 37\%	& 15\%	& 30\%	& 85\%	& 63\%	& 29\%	& 45\%	& 9\%	& 56\%	\\
\textbf{\% in Cluster 1}	& 15\%	& 63\%	& 85\%	& 70\%	& 15\%	& 37\%	& 71\%	& 55\%	& 91\%	& 44\%	\\
\textbf{Silhouette Score}	& 0.16	& 0.08	& 0.15	& 0.11	& 0.13	& 0.11	& 0.09	& 0.09	& 0.3	& 0.07	\\
\bottomrule
\end{tabular}
\end{table*}

\textbf{Exclusionary Reclassification}: The top two rows of Table \ref{tab:cluster_analysis} show how the excluded cluster was classified by the newly trained model. The third row shows the ExRe score for poisoned and clean clusters. Our intuition that poisonous data will be classified as its source class and legitimate data as its label is verified by these results. Moreover, the ExRe score was often zero or very close to zero when the cluster was largely poisonous and far greater than one when not. Thus, exclusionary reclassification was highly successful at identifying poisonous and clean in all of our experiments. Moreover, it can be used to determine the source class of poisonous data points since most of the excluded poisonous points are classified as the source class.

\textbf{Relative Size}: Rows 4 and 5 of Table \ref{tab:cluster_analysis} shows how much of the data was placed in each cluster for clean and poisoned classes. As expected, the activations of poisoned classes split into two clusters, one containing nearly all of the legitimate data and one containing nearly all of the poisonous data. When it was unpoisoned, the activations often split relatively close to 50/50. In the worst cases (MNIST 4 and 7+), we saw a 70/30 split. 

\textbf{Silhouette Score}: Row 6 of Table \ref{tab:cluster_analysis} shows the silhouette score for clean and poisoned classes. Most of the poisoned classes tend to have a silhouette score of at least .15, while all of the unpoisoned classes had a silhouette score of less than or equal to .11. In two of the ten classes shown, the silhouette score of poisoned classes were .10 and .09. Therefore, a threshold between .10 and .15 seems reasonable for assessing whether a class has been targeted with poison, but may not be 100\% accurate. Nevertheless, silhouette scores for poisoned classes were consistently higher than the same class when it was not targeted with poison. Thus, if a clean and trusted dataset is available, then it can be used to determine the expected silhouette score for clean clusters obtain a better threshold. 
We also experimented with using the gap statistic \citep{tibshirani2001estimating} to compare the relative fit of one versus two clusters but this was largely unsuccessful.

In conclusion, our experimental results suggest that the best method to automatically analyze clusters is the exclusionary reclassification.

\subsection{Backdoor Repair}
Finally, we evaluated the proposed backdoor repair technique on the base poisonous MNIST model and found that re-training converged after 14 epochs on the repaired samples only. In contrast, re-training from scratch took 80 epochs on the full data set. The error rates prior to and after re-training of a model that was trained on a 33\% poisoned dataset are shown in Table \ref{tab:retrain}. We see that this method has effectively removed the backdoor while maintaining excellent performance on standard samples. Our results suggest this is an effective way to remove backdoors.

\begin{table}[htb!]
  \centering
  \caption{Test error for different classes across poison and clean data prior to and after re-training.}\label{tab:retrain}
  \footnotesize
  \begin{tabular}{ c c c c c c c c c c c}
  \toprule
    Class & 0 & 1 & 2 & 3 & 4 \\
    \hline
    Before (\%) &  35.19    & 32.62   & 33.47   & 31.96   & 33.2 \\
    After (\%) &  0.96    & 0.71  & 0.6   & 0 & 0.21  \\
    \hline
    Class & 5& 6 & 7 & 8 & 9 \\
    \hline
    Before (\%) & 32.34   & 34.68   & 35.08   & 33.61   & 31.29 \\
    After (\%) & 1.6   & 0   & 1.36  & 0.82  & 0.61 \\
  \bottomrule
  \end{tabular}
\end{table}

\section{Discussion}
\label{sec:conclusion}

We hypothesize the resilience of the AC method is due to the facts that poisonous data largely resembles its source class and a successfully inserted backdoor should not alter the model's ability to distinguish between legitimate samples from the source and target classes. Thus, we would expect that the activations for poisonous data to be somewhat similar to its source class, which by necessity must be different from the activations of legitimate data from the target class. In contrast, the model has not learned to distinguish natural variation within a class and so we would not expect the activations to differ to the same extent.

\begin{figure}[htb]
\centering
\includegraphics[width=.4\textwidth]{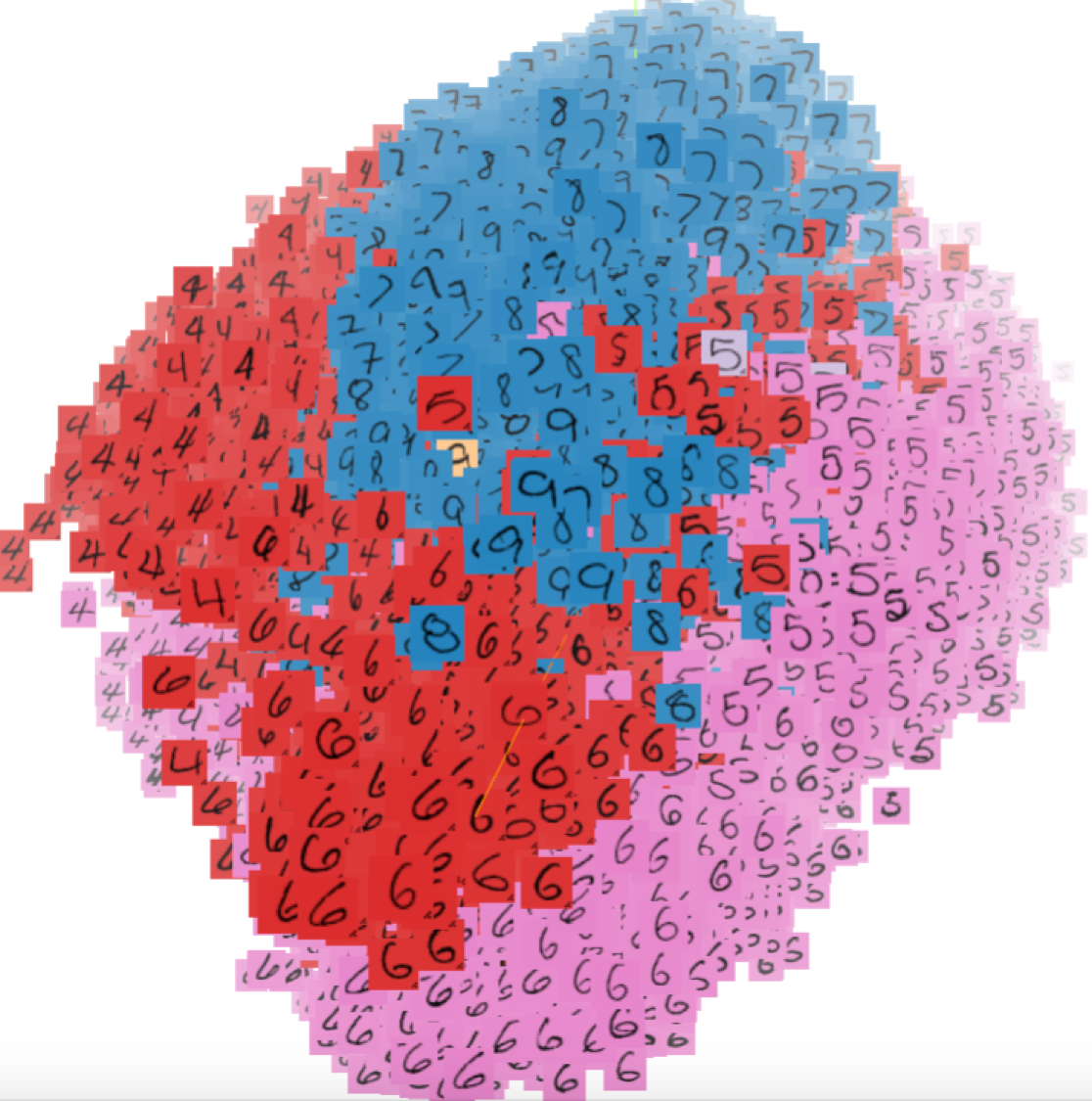}
\caption{Activations of the 7+ class, which has been poisoned with data sourced from the 4, 5, and 6 classes, shown together with activations of legitimate data from the 4, 5, and 6 classes projected onto the first 3 principle components. True positives are shown in red, true negatives in blue, false positives in yellow, and legitimate data from the poison source classes in purple.}
\label{fig:7+}
\end{figure}

This hypothesis is supported by Figure \ref{fig:7+}. Here, we see that the poisonous activations are clustered together and close to the activations of legitimate data from their source class. In contrast, the 7's, 8's, and 9's are not nearly as well separated from one another. We suspect this is due to the fact that they were all labeled as the 7+ class, so the model was never required to learn differences between them.

An adversary may attempt to circumvent our defense. However, this poses unique challenges. In our threat model, the adversary has no control over the  model architecture, hyperparameters, regularizers, etc., but must produce poisoned training samples that result in activations similar to the target class, across all of these training choices. Standard techniques for generating adversarial samples could be applied to ensure that poisonous data activate similarly to the target class, but there is no guarantee that the backdoor would generalize to new samples and the model would likely overfit \citep{ZhangBHRV16}. Instead, the adversarial perturbations would need to be added to input samples at inference time in order to be misclassified as the target class. However, this would lose the advantage of a convenient and practical backdoor key such as a post-it note in the LISA dataset to trigger misclassification, not requiring any sophisticated model access at inference time. Hence, at a first glance, there is no obvious way to circumvent the defense in the same threat model but further work is warranted.

\section{Conclusion}

Vulnerabilities of machine learning models pose a significant safety risk. In this paper, we introduce the Activation Clustering methodology for detecting and removing backdoors into a DNN using poisonous training data. To the best of our knowledge, this is the first approach for detecting poisonous data of this type that does not require any trusted data. Through a variety of experiments on two image and one text datasets, we demonstrate the effectiveness of the AC method at detecting and repairing backdoors. Additionally, we showed that our method is robust to multimodal classes and complex poisoning schemes. We implemented and released our method through the open source IBM Adversarial Robustness Toolbox \citep{nicolae2018adversarial}. Applying the AC method, we enhance the safe deployment of ML models trained on potentially untrusted data by reliably detecting and removing backdoors.

\clearpage{}
\balance
\bibliographystyle{apalike}
\bibliography{bib}

\begin{thebibliography}{}

\bibitem[Aggarwal et~al., 2001]{aggarwal2001surprising}
Aggarwal, C.~C., Hinneburg, A., and Keim, D.~A. (2001).
\newblock On the surprising behavior of distance metrics in high dimensional
  space.
\newblock In {\em International conference on database theory}, pages 420--434.
  Springer.

\bibitem[Baracaldo et~al., 2017]{baracaldo2017mitigating}
Baracaldo, N., Chen, B., Ludwig, H., and Safavi, J.~A. (2017).
\newblock Mitigating poisoning attacks on machine learning models: A data
  provenance based approach.
\newblock In {\em Proceedings of the 10th ACM Workshop on Artificial
  Intelligence and Security}, pages 103--110. ACM.

\bibitem[Barreno et~al., 2010]{survey:adv-ml-barreno2010}
Barreno, M., Nelson, B., Joseph, A.~D., and Tygar, J. (2010).
\newblock The security of machine learning.
\newblock {\em Machine Learning}, 81(2):121--148.

\bibitem[Britz, 2015]{britz2015text}
Britz, D. (2015).
\newblock Implementing a cnn for text classification in tensorflow.
\newblock \url{https://ibm.biz/BdYwmY}.

\bibitem[Carlini and Wagner, 2017]{carlini2017towards}
Carlini, N. and Wagner, D. (2017).
\newblock Towards evaluating the robustness of neural networks.
\newblock In {\em 2017 IEEE Symposium on Security and Privacy (SP)}, pages
  39--57. IEEE.

\bibitem[Domingos, 2012]{domingos2012few}
Domingos, P. (2012).
\newblock A few useful things to know about machine learning.
\newblock {\em Communications of the ACM}, 55(10):78--87.

\bibitem[Gu et~al., 2017]{badnets}
Gu, T., Dolan{-}Gavitt, B., and Garg, S. (2017).
\newblock Badnets: Identifying vulnerabilities in the machine learning model
  supply chain.
\newblock {\em CoRR}, abs/1708.06733.

\bibitem[Huang et~al., 2011]{training_poisoning}
Huang, L., Joseph, A.~D., Nelson, B., Rubinstein, B.~I., and Tygar, J.~D.
  (2011).
\newblock Adversarial machine learning.
\newblock In {\em Proceedings of the 4th ACM Workshop on Security and
  Artificial Intelligence}, AISec '11, pages 43--58, New York, NY, USA. ACM.

\bibitem[Kim, 2014]{kim2014convolutional}
Kim, Y. (2014).
\newblock Convolutional neural networks for sentence classification.
\newblock {\em arXiv preprint arXiv:1408.5882}.

\bibitem[Kloft and Laskov, 2010]{kloft2010online}
Kloft, M. and Laskov, P. (2010).
\newblock Online anomaly detection under adversarial impact.
\newblock In {\em Proceedings of the Thirteenth International Conference on
  Artificial Intelligence and Statistics}, pages 405--412.

\bibitem[Kloft and Laskov, 2012]{kloft2012security}
Kloft, M. and Laskov, P. (2012).
\newblock Security analysis of online centroid anomaly detection.
\newblock {\em Journal of Machine Learning Research}, 13(Dec):3681--3724.

\bibitem[Liu et~al., 2018]{liu2018fine}
Liu, K., Dolan-Gavitt, B., and Garg, S. (2018).
\newblock Fine-pruning: Defending against backdooring attacks on deep neural
  networks.
\newblock {\em arXiv preprint arXiv:1805.12185}.

\bibitem[Liu et~al., 2017a]{liu2017trojaning}
Liu, Y., Ma, S., Aafer, Y., Lee, W.-C., Zhai, J., Wang, W., and Zhang, X.
  (2017a).
\newblock Trojaning attack on neural networks.

\bibitem[Liu et~al., 2017b]{liu2017neural}
Liu, Y., Xie, Y., and Srivastava, A. (2017b).
\newblock Neural trojans.
\newblock In {\em Computer Design (ICCD), 2017 IEEE International Conference
  on}, pages 45--48. IEEE.

\bibitem[Mu{\~n}oz-Gonz{\'a}lez et~al., 2017]{munoz2017towards}
Mu{\~n}oz-Gonz{\'a}lez, L., Biggio, B., Demontis, A., Paudice, A.,
  Wongrassamee, V., Lupu, E.~C., and Roli, F. (2017).
\newblock Towards poisoning of deep learning algorithms with back-gradient
  optimization.
\newblock {\em arXiv preprint arXiv:1708.08689}.

\bibitem[Nelson et~al., 2009]{original-roni}
Nelson, B., Barreno, M., Chi, F.~J., Joseph, A.~D., Rubinstein, B.~I., Saini,
  U., Sutton, C., Tygar, J., and Xia, K. (2009).
\newblock Misleading learners: Co-opting your spam filter.
\newblock In {\em Machine learning in cyber trust}, pages 17--51. Springer.

\bibitem[Nelson, 2010]{ch-roni-nelson2010}
Nelson, B.~A. (2010).
\newblock {\em Behavior of Machine Learning Algorithms in Adversarial
  Environments}.
\newblock University of California, Berkeley.

\bibitem[Nicolae et~al., 2018]{nicolae2018adversarial}
Nicolae, M.-I., Sinn, M., Tran, M.~N., Rawat, A., Wistuba, M., Zantedeschi, V.,
  Molloy, I.~M., and Edwards, B. (2018).
\newblock Adversarial robustness toolbox v0. 2.2.
\newblock {\em arXiv preprint arXiv:1807.01069}.

\bibitem[Papernot et~al., 2016]{papernot2016survey}
Papernot, N., McDaniel, P., Sinha, A., and Wellman, M. (2016).
\newblock Towards the science of security and privacy in machine learning.
\newblock {\em arXiv preprint arXiv:1611.03814}.

\bibitem[Simonyan and Zisserman, 2014]{simonyan2014very}
Simonyan, K. and Zisserman, A. (2014).
\newblock Very deep convolutional networks for large-scale image recognition.
\newblock {\em arXiv preprint arXiv:1409.1556}.

\bibitem[Steinhardt et~al., 2017]{steinhardt2017certified}
Steinhardt, J., Koh, P. W.~W., and Liang, P.~S. (2017).
\newblock Certified defenses for data poisoning attacks.
\newblock In {\em Advances in Neural Information Processing Systems}, pages
  3517--3529.

\bibitem[Tibshirani et~al., 2001]{tibshirani2001estimating}
Tibshirani, R., Walther, G., and Hastie, T. (2001).
\newblock Estimating the number of clusters in a data set via the gap
  statistic.
\newblock {\em Journal of the Royal Statistical Society: Series B (Statistical
  Methodology)}, 63(2):411--423.

\bibitem[Yang et~al., 2017]{yang2017generative}
Yang, C., Wu, Q., Li, H., and Chen, Y. (2017).
\newblock Generative poisoning attack method against neural networks.
\newblock {\em arXiv preprint arXiv:1703.01340}.

\bibitem[Zhang et~al., 2016]{ZhangBHRV16}
Zhang, C., Bengio, S., Hardt, M., Recht, B., and Vinyals, O. (2016).
\newblock Understanding deep learning requires rethinking generalization.
\newblock {\em CoRR}, abs/1611.03530.

\end{thebibliography}

\clearpage
\appendix

\end{document}